%% file: main.tex
\documentclass[letterpaper]{article}
\usepackage[hyphens]{url}
\input{figures.tex}

\title{Completion Reasoning Emulation for the Description Logic \ELp}
\author{Aaron Eberhart, Monireh Ebrahimi, Lu Zhou, Cogan Shimizu, and Pascal Hitzler\\
Data Semantics Lab\\
Kansas State University, USA\\
\{aaroneberhart,monireh,luzhou,coganmshimizu,hitzler\}@ksu.edu
}

\begin{document}

\maketitle

\begin{abstract}
We present a new approach to integrating deep learning with knowledge-based systems that we believe shows promise. Our approach seeks to emulate reasoning structure, which can be inspected part-way through, rather than simply learning reasoner answers, which is typical in many of the black-box systems currently in use. We demonstrate that this idea is feasible by training a long short-term memory (LSTM) artificial neural network to learn \ELp\ reasoning patterns with two different data sets. We also show that this trained system is resistant to noise by corrupting a percentage of the test data and comparing the reasoner's and LSTM's predictions on corrupt data with correct answers.
\end{abstract}

\sect{Introduction}
Machine learning and neural network techniques are often contrasted with logic-based systems as opposite in many regards. The challenge of integrating inductive learning strategies with deductive logic has no easy or obvious solution. Though there are doubtless many reasons for this, one major roadblock is that solutions and evaluations which work well for logic, or for deep learning and machine learning, do not work well in the opposite, and likely do not further the goal of integration. In this paper we present a new approach that embraces the liminality of this specific integration task. Our approach is by its very nature ill-suited to either machine learning or logic alone. But it tries to avoid the pitfalls of unintentionally favoring one paradigm over the other, aiming instead to grasp at something new in the space between.

\subsect{Related Work}
A popular approach for training neural networks to recognize logic is to use embedding techniques borrowed from the field of natural language processing (NLP). For instance, Makni and Hendler designed a system that layers Resource Description Framework (RDF) graphs into adjacency matrices and embedding them \cite{makni2018deep}. There is also considerable research into path-finding in embedded triple data using recurrent neural networks (RNNs) or LSTMs \cite{xiong2017deeppath,DBLP:journals/corr/DasNBM16,das2017go}. Memory networks also successfully operate over triple embeddings and are capable of transferring training to work over triple embeddings from completely different vocabularies \cite{ebrahimi2018reasoning}. There is even an attempt to embed \ELpp\ with TransE using a novel concept of $n$-balls, though it currently does not consider the RBox as well as certain \ELpp\ axioms that do not translate into a triple format\cite{kulmanov2019embeddings}\cite{bordes2013translating}.

\subsect{Discussion}
Many of the systems just mentioned are very successful, yet have a tendency to adopt familiar evaluation strategies from deep learning. Sure, it's great to be able to get a 99\% F1-score. When we do not know the underlying structure of the data that we are using, this is often the best we can hope for. However, precision and recall do not make much sense in a knowledge base evaluation. For an integrated system, some new goals and evaluation strategies seem warranted that are neither completely deductive nor entirely empirical. 

In a deductive reasoning system the semantics of the data is known {\em explicitly}. Why then would we want to blindly allow such a system to teach itself what is important when we already know what and how it should learn? Possibly we don't know the best ways to guide a complex network to the correct conclusions, but surely more, not less, transparency is needed for integrating logic and deep learning. Transparency often becomes difficult when we use extremely deep or complex networks that cannot be reduced to components. It also makes things difficult when we pre-process our data to help the system train by doing embeddings or other distance adjustments. When we embed all the similar things close together and all the dissimilar things far apart the system can appear to succeed, but we can't tell if it was the embedding or the system itself or one of a dozen other things that might have caused this. 

In response to these and other concerns we have developed some research questions that we hope to address in this paper. Though we certainly will not be able to definitively answer all of them, we do hope that by starting this investigation we can demonstrate future potential for this type of inquiry.

\subsect{Research Questions}
\goals
Question one is the primary focus of this project, and the one to which all the others are subordinate. It is the intention of this paper and the prototype system we have implemented to demonstrate that this is a feasible goal. Whatever disagreements may arise concerning our later investigations, we will argue that this is a necessary and plausible alternative course of research.

The second question we have already mentioned in the last section. It is our intention to avoid any pre-processing that will allow the system to give correct answers without learning the reasoning patterns. In this way we can better verify whether we have truly solved the reasoning problem and not some other easier task.

The third question is related  to the second, but imposes a more broad restriction on what we will be allowed to do. Not only do we want to avoid embedding the data in a way that will help the system to learn the answers. Also we must take care when we go from logic to numerical data, and back again, not to interfere with the latent semantics in the logic. By avoiding distance comparisons in the naming convention for the input we ensure that the names are arbitrary in both systems. The only distances we intend to measure are between predictions and answers.

The fourth question is very important for our conception of how future systems should work. In a logic-based system there is transparency at any level of reasoning. We cannot, of course, expect this in most neural networks. With a network that aims to emulate reasoner behavior, however, we can impose a degree of intermediate structure. This intermediate structure allows us to inspect a network part-way through and perform a sort of "debugging" since we know exactly what it should have learned at that point. This is a crucial break from current thinking that advocates more and deeper opaque hidden layers in networks that improve accuracy but detract from explainability. Inspection of intermediate answers could indicate whether a proposed architecture is actually learning what we intend, which should aid development of more correct systems.

The fifth and final question is a difficult but necessary question, likely beyond the scope of this paper. For now, it is primarily a reminder that the method we choose intentionally does not try to find best answers for individual statements. It tries to emulate reasoner behavior by matching the shape of the reasoning patterns. That is not to say we don't care about right answers, or that they don't matter. Those values are reported for our system. And if reasoning structure is matched well enough then correct answers should follow. But we believe that the reasoning distance results we report are far more compelling and tell a better story about what is happening, as well as indicate the future potential of this method.

\sect{Approach}
We present a method for learning the structure of \ELp\ reasoning patterns using an LSTM that tests the questions posed above. In our system we avoid, as much as possible, embeddings or any distance adjustments that might help the system learn answers based on embedding similarity without first learning the reasoning structure. In fact, a primary feature of our network is its lack of complexity. We take a very simple logic, reason over knowledge bases in that logic, and then extract supports from the reasoning steps, mapping the reasoner supports back to sets of the original knowledge base axioms. This allows us to encode the input data in terms of only knowledge base statements. It also provides an intermediate answer that might improve results when provided to the system. This logic data is fed into three different LSTM architectures with identical input and output dimensionalities. One architecture, which we call "Deep", does not train with support data but has a hidden layer the same size as the supports we have defined. Another architecture, called "Piecewise", trains two separate half-systems, one with knowledge bases and one with supports from the reasoner. The last system, called "Flat", simply learns to map inputs directly to outputs for each reasoning step. We will discuss in detail each part of the system in the following sections, then provide some results.

\subsection{\ELp}\label{subsubsec:EL+}
\ELp\ is a lightweight and highly tractable description logic. A typical reasoning task in \ELp\ is a sequential process with a fixed endpoint, making it a perfect candidate for sequence learning. Unlike RDF, which reasons over instance data in triple format, \ELp\ reasoning occurs on the predicate level. Thus we will have to train the system to actually learn the reasoning patterns and logical structure of \ELp\ directly from encoded knowledge bases.

\subsubsect{Syntax}
\elpSignature \elpSyntax

\subsubsect{Semantics}
\elpSemantics An interpretation $I$ where $I=(\Delta^{I},\cdot^{I})$ maps $N_I, N_C, N_R$ to elements, sets, and relations in $\Delta^I$ with function $\cdot^I$. For an interpretation $I$ and $C(i), R(i) \in \Sigma$, the function $\cdot^{I}$ is defined in Table \ref{tab:sem}. 

\subsubsect{Reasoning}
It is a well established result that any \ELp\ knowledge base has a least fixed point that can be determined by repeatedly applying a finite set of axioms that produce all entailments of a desired type \cite{DBLP:journals/corr/abs-1711-03902}\cite{kazakov2012elk}.\completionRules In other words, we can say that reasoning in \ELp\ often amounts to a sequence of applications of a set of pattern-matching rules. One such set of rules, the set we have used in our experiment, is given in Table \ref{tab:comprul}. The reasoning reaches {\em completion} when there are no new conclusions to be made. Because people are usually interested most in concept inclusions and restrictions, those are the types of statements we choose to include in our reasoning.

\suppExample After the reasoning has finished we are able to recursively define supports for each conclusion the reasoner reaches. The first step, of course, only has supports from the knowledge base. After this step supports are determined by effectively running the reasoner in reverse, and replacing each statement that is not in the original knowledge base with a superset that is, as you can see by the colored substitutions in Table \ref{tab:support}. When the reasoner proved the last statement it did not consider all of the supports, since it had already proved them. It used the new facts it had learned in the last iteration but we have drawn their supports back out so that we can define a fixed set of inputs from the knowledge base.

\subsect{Synthetic Data}
To provide sufficient training input to our system we provide a synthetic generation procedure that combines a structured forced-lower-bound reasoning sequence with a connected randomized knowledge base. This allows us to rapidly generate many normal semi-random \ELp\ knowledge bases of arbitrary reasoning difficulty. For this experiment we choose a moderate difficulty setting so that it can compare with non-synthetic data.  An example of one iteration of the two-part sequence is provided in Figure \ref{fig:seq}. To ensure that the randomized statements do not shortcut this pattern, the random statements are generated in a nearly disjoint space and connected only to the initial seed term. This ensures that at least one element of the random space will also produce random entailments for the duration of the sequence, possibly longer. Our procedure also guarantees that each completion rule will be used at least once every iteration of the sequence so that all reasoning patterns can potentially be learned by the system.
\oneSequence

\subsect{SNOMED}
We have also imported data from the SNOMED 2012 ontology to ensure that our method is applicable to non-synthetic data. SNOMED is a widely-used, publicly available, ontology of medical terms and relationships.\cite{DeSilva:2011:SNM:1950978.1951053} SNOMED 2012 has 39392 logical axioms, some of which are complex, but this can be normalized in constant time to a logically equivalent set of 124,428 axioms. It is expressed in \ELpp, and newer versions utilize this complexity more liberally. But the 2012 version contains exactly one statement that is not in \ELp\ after normalization. We feel confident that omitting $\top \sqsubseteq \exists$topPartOf.Self does not constitute a substantial loss in the ontology. 

Because SNOMED is so large, we sample it to obtain connected subsets that also produce a minimum number of reasoning steps. We require that the samples be connected because any normal knowledge base is connected, and it improves the chances that the statements will have entailments. Often, however, random connected samples do not entail anything, and this is also unusual in a normal ontology, so we require a minimum amount of reasoning activity in the samples as well. The reasoning task for SNOMED is more unbalanced than for the synthetic data. It is equally trivial for the reasoner to solve either knowledge base type. However, we observe that random connected sampling tends to favor application of rules 3, 5, and 6 much more heavily than others so the system will have a more difficult time learning the overall reasoning patterns. This imbalance is likely an artifact from SNOMED because it seems to recur in different sample sizes with different settings, though we acknowledge that it could be correlated somehow with the sample generation procedure.

\subsect{LSTM}
We choose to use an LSTM to learn the reasoning patterns of \ELp\ because LSTMs are designed to learn sequences. The network is kept as simple as possible to achieve the outputs we require and maximize transparency. \piecewisePic For the piecewise system, depicted in Figure \ref{fig:piece}, we use two separate flat single LSTMs that have the number of neurons matching first the support shape, and later the output shape. \deepPic The deep system shown in Figure \ref{fig:deep} is constructed to match this shape so that the two will be comparable, it simply stacks the LSTMs together so that it can train without supports. \flatPic We also train a flat system, shown in Figure \ref{fig:flat}, that has the same input and output shape but lacks the internal supports. We have done this three ways because we can compare the behavior of the systems with support against the flat system as a baseline, and then see whether the piecewise training is helping. All designs treat the reasoning steps as the sequence to learn, so the number of LSTM cells is defined by the maximum reasoning sequence length. We train our LSTMs using regression on mean squared error to minimize the distances between predicted answers and correct answers.

\subsect{Overall System}
The individual parts of our system are not themselves novel. Our result, however, lies rather in the unexpected success of this unconventional general approach we have applied. One of the crucial factors in making this work is the way in which we have translated the data from its logical format into something a neural network can understand. In the following sections we will discuss how our system transforms data from a knowledge base format to an LSTM format and back again.

\subsubsect{Statement Encoding}
 Every data set we use has a fixed number of names that it can contain for both roles and concepts, and every concept or role has an arbitrary number name identifier. For the synthetic data these numbers are randomly generated. For SNOMED data, all of the labels are stripped and the concept and role numbers are shuffled around and substituted with random integers to remove any possibility of the names injecting a learning bias. These labels are remembered for later retrieval for output but the reasoner and LSTM do not see them. It can seem occasionally like there is a bias in the naming if output for SNOMED data is inspected because all the labels seem related. This is merely a result of our sampling technique which requires connected statements, so it is not a concern. 
 
 Knowledge bases have a maximum number of role and concept names that are used to scale all of the integer values for names into a range of $[-1,1]$. To enforce disjointness of concepts and roles, we map all concepts to $(0,1]$, all roles to $[-1,0)$.\statementEncodingTable Each of the six possible normalized axiom forms is encoded into a 4-tuple based on the logical encodings defined in Table \ref{tab:encode}. 
 
 We would like to emphasize the distinction between our method, which we prefer to call an encoding, and a traditional embedding. This encoding adds no additional information to the names beyond the transformation from integer to scaled floating point number. It can be reversed, with a slight loss of precision due to integer conversion, quite directly. The semantics of each encoded predicate name with regards to its respective knowledge base is structural in relation to its position in the 4-tuple, just like it would be in an \ELp\ axiom. And its semantics is not at all related to the other non-equal predicate names within the same 4-tuple.

\subsubsect{Knowledge Base Encoding}
In order to input knowledge bases into the LSTM we first apply the encoding defined in the previous section to each of its statements. Then we concatenate each of these encodings end-to-end to obtain a much longer vector. For instance, $$\text{C}2 \sqsubseteq \text{C}1, \text{C}3 \sqsubseteq \text{C}4, \text{C}4 \sqsubseteq \exists \text{R}1.\text{C}2$$ might translate to $$[0.0,0.5,0.25,0.0,0.0,0.75,1,0.0,0.0,1.0,-1.0,0.5].$$ This knowledge base vector is then copied the same number of times as there are reasoning steps and stuck together once again along a new dimension. For an experiment this is done hundreds or thousands of times, and these fill up the tensor that is given to the LSTM to learn. Because some of the randomness can cause ragged data, any tensor dimensions that are not the same length as the maximum are padded with zeros.

\sect{Results}
Viewed in terms of the questions posed at the beginning of this paper we feel that our results are solid, and they validate a reexamination of how we should approach integrating logic and neural networks. We have attempted to create a system that emulates a reasoning task rather than reasoning output.

\exampleSynPred If we examine the example output from the synthetic data inputs in Table \ref{tab:synex}, it is clear that it is getting very close to many correct answers. When it misses, it still appears to be learning the shape, and this makes us optimistic about its future potential. \exampleSnoPred The SNOMED predictions are much more dense and do not fit well into a table, but we have included a few good examples with the original data labels translated into English sentences. Because there are so many near-misses we include edit distance evaluations that better capture the degree to which each predicted statement misses what it should have been. 

\accTable \distTable It is interesting to note that by comparing Table \ref{tab:fscore} with Table \ref{tab:dists} we can see that on the much harder SNOMED data the deep system {\em appears} to have a better result because of the higher F1 score, but the average edit distance, which is our preferred alternative measure for evaluation, is not obviously correlated with the F1-score. This confirms that our fifth research question was appropriate and that F1-score might not be sufficient to evaluate whether or not we can learn the shape of reasoning patterns. 

\syntrainingPic \snotrainingPic A cause for the discrepancy may be the higher training difficulty for reaching the completion versus reaching the supports in the SNOMED data, which you can see in Figures \ref{fig:syntrain} and \ref{fig:snotrain}. We can speculate on the degree to which various factors are contributing to this, but importantly, the fact that this discussion is even possible in the first place is a confirmation that we can answer the fourth research question in the affirmative. 

Another interesting result can be seen by examining the charts in Figures \ref{fig:distlev} - \ref{fig:fpred}. As we have noticed before, the Deep architecture performs quite well on the synthetic data and is quite resistant to increased levels of noise. The flat system gets similar results, but slightly worse. Surprisingly, the piecewise model starts out mostly the same as the deep system on synthetic data but then very quickly falters. When we examine the results for the harder SNOMED data, however, the opposite seems to be happening. The deep and flat systems still track each other. But for the unbalanced data the deep system crosses into the random range by 30-50\% corruption, while the piecewise system is able to skirt just above random for the Character and Predicate distance functions until around 80-90\%.  

\subsect{Methodology}
 Our system is trained using randomized 10-fold cross validation to 20000 epochs on the deep and flat systems and 10000 epochs each for the parts of the piecewise system at a learning rate of 0.0001. The results from every fold are saved and we report the average of all of the runs. Often there are better and worse runs in any given 10-fold validation but the averages across multiple cross-validations with the same settings remain basically the same. 10-fold cross-validations are performed with an extra comparison against data that has been corrupted at a fixed rate, starting with zero percent probability of corruption and moving upwards by ten percent each time. The data in Tables \ref{tab:fscore} and \ref{tab:dists} is from the comparison with no corruption, so the error data is not reported (it is identical with reasoner answers).
 
 To perform our evaluations we use three unique distance measurements. We have a naive "Character" Levenshtein distance function that takes two unaltered knowledge base statement strings and computes their edit distance.\cite{lev:2019} However, because some names in the namespace are one digit numbers and other names are two digit numbers, we include a modified version of this function, called "Atomic", that uniformly substitutes all two digit numbers in the strings with symbols that do not occur in either. Since there cannot be more than eight unique numbers in two decoded strings there are no issues with finding enough new symbols. By doing the substitutions we can see the impact that the number digits were having on the edits from the Atomic Levenshtein distance. Finally we devise a distance function that is based on our encoding scheme. The Predicate Distance method disassembles each string into only its predicates. Then, for each position in the 4-tuple, a distance is calculated that yields zero for perfect matches, absolute value of guessed number - actual number for correct Class and Role guesses, and guessed number + actual number for incorrect Class and Role matches. So, for instance, guessing C1 when the answer is C2 will yield a Predicate Distance of 1, while a guess of R2 for a correct answer of C15 will yield 17. Though this method is specific to our unique encoding, we believe it detects good and bad results quite well because perfect hits are 0, close misses are penalized a little, and large misses are penalized a lot. 
 
 For each method we take every statement in a knowledge base completion and compare it with the best match in the reasoner answer, random answers, and corrupted knowledge base answers. While we compute these distances we are able to obtain precision, recall, and F1-score by counting the the number of times the distance returns zero and treating the statement predictions as classifications. Each time the system runs it can make any number of predictions, from zero to the maximum size of the output tensor. This means that, although the predictions and reasoner are usually around the same size and those comparisons are fine, we have to generate random data to compare against that is as big as could conceivably be needed by the system. Any artificial shaping decisions we made to compensate for the variations between runs would invariably introduce their own bias in how we selected them. Thus the need to use the biggest possible random data to compare against means the precision, recall, and F1-score for random are low.

\sect{Conclusion}
In this experiment demonstrate that \ELp\ reasoning can be emulated with an LSTM. Although our example is merely a prototype, we can definitively say yes to the first research question, a neural network can be trained to emulate completion reasoning behavior. And because we have answered this first research question without using name adjustments or embeddings, so we can also answer yes to the second and third questions. We have already discussed questions four and five. 

\subsect{Future Work}
There are a number of potential improvements that could build on the work presented here. A few of the design choices we made might have had an impact of the results of the study and we are curious if alternate strategies could improve the way that the system learns reasoning patterns. We might have chosen a 3-tuple encoding pattern that was more dense but did not mirror the shape of the statements symmetrically. We also considered adding the statements as an additional dimension rather than concatenating them all together. This may have made it easier for the system to distinguish individual statements but would also add a lot of complexity in the extra dimension. We experimented with different neuron types like gated recurrent unit (GRU) and basic RNN, though these had no significant impact on the results. The type of neuron seems to have less effect than the overall structure of the network. It would be interesting to see if different ways of segmenting a network, besides just along the supports as we have defined them, could improve results.

Along with these ideas to improve our current system, we also speculate that with some effort we should be able to accomplish basically the same type of thing with a more expressive logic and a new encoding. \ELpp\ for instance would be an easy first attempt since it is so similar to \ELp. Though we should be able to adapt this strategy for any logic that uses completion rules for reasoning. Whatever we choose to do with it, we are optimistic that we can use this strategy to advance understanding of how logic and neural networks can work together.

\subsect{Testing Environment}
All testing was done on a computer running Ubuntu 19.10 64-bit with an Intel Core i7-9700K CPU@3.60GHz x 8, 47.1 GiB DDR4, and a GeForce GTX 1060 6GB/PCIe/SSE2. Source code and experiment data is available on GitHub \url{https://github.com/aaronEberhart/PySynGenReas}.

\bibliographystyle{aaai}
\bibliography{refs}

\levDist\levAcc\atomDist\atomAcc\predDist\predAcc 

\end{document}

%% file: figures.tex
\usepackage{xcolor,graphicx,subfig,url,amsmath,aaai20,times,helvet,courier,multicol,mdframed}

\input{macros.tex}

\def\statementEncodingTable{
\def\arraystretch{1.65}
\setlength\tabcolsep{8pt}
\begin{table}
	\caption{Translation Rules}
	\label{tab:encode}	
	\begin{center}
		\makebox[\linewidth]{
			\begin{tabular}{|ccc|}
				\hline
				KB statement&\vline&Vectorization\\
				\hline
				CX $\sqsubseteq$ CY&$\rightarrow$&$[\ 0.0,\ \frac{X}{c}, \frac{Y}{c},\ 0.0\ ]$\\
				\hline
				CX $\sqcap$ CY $\sqsubseteq$ CZ&$\rightarrow$&$[\ \frac{X}{c},\ \frac{Y}{c},\ \frac{Z}{c},\ 0.0\ ]$\\
				\hline
				CX $\sqsubseteq$ $\exists$RY.CZ&$\rightarrow$&$[0.0,\ \frac{X}{c},\ \frac{-Y}{r},\ \frac{Z}{c}]$\\
				\hline
				$\exists$RX.CY $\sqsubseteq$ CZ&$\rightarrow$&$[\frac{-X}{r},\ \frac{Y}{c},\ \frac{Z}{c},\ 0.0]$\\
				\hline
				RX $\sqsubseteq$ RY&$\rightarrow$&$[0.0, \frac{-X}{r}, \frac{-Y}{r}, 0.0]$\\
				\hline
				RX $\circ$ RY $\sqsubseteq$ RZ&$\rightarrow$&$[\frac{-X}{r}, \frac{-Y}{r}, \frac{-Z}{r}, 0.0]$\\
				\hline 
			\end{tabular}		
	}
	
	\medskip
	c = Number of Possible Concept Names \\ r = Number of Possible Role Names	
	\end{center}
\end{table}
\def\arraystretch{1}
\setlength\tabcolsep{6pt}
}

\def\elpSyntax{
Expressions in \ELp\ use the following grammar:\nls
\def\arraystretch{1.35}
\begin{tabular}{lcccccc}\label{tab:syn}
\centering
$\textbf{R}$&$::=$&$N_R$&$|$&$\textbf{R} \circ \textbf{R}$\\
$\textbf{C} $&$::=$&$ N_C $&$|$&$ \textbf{C} \sqcap \textbf{C} $&$|$&$ \exists \textbf{R}.\textbf{C}$
\end{tabular}\def\arraystretch{1}
}

\def\elpSignature{
The signature $\Sigma$ for \ELp\ is defined as:\nls
$\Sigma = \left<N_I,N_C,N_R\right>\ $ with $N_I, N_C, N_R$ pairwise disjoint.\nls\noindent
$N_I$ is a set of individual names.\\
$N_C$ is a set of Concept names that includes $\top$.\\
$N_R$ is a set of Role names.\nls
}

\def\elpSemantics{
\def\arraystretch{1.15}
\setlength\tabcolsep{8pt}
\begin{table*}
	\caption{\ELp\ Semantics}
	\label{tab:sem}
	\begin{center}
		\makebox[\linewidth]{
			\begin{tabular}{ |c|c|c| } 
				\hline
				Description & Expression & Semantics\\ 
				\hline 
				Individual & $ a $ & $ a \in \Delta^{\mathcal{I}} $\\    
				Top & $ \top $ & $ \Delta^{\mathcal{I}} $\\ 
				Bottom & $ \bot $ & $ \emptyset $ \\
				Concept & $ C $ & $ C^{\mathcal{I}} \subseteq \Delta^{\mathcal{I}} $\\
				Role & $ R $ & $ R^{\mathcal{I}} \subseteq \Delta^{\mathcal{I}} \times \Delta^{\mathcal{I}} $\\
				Conjunction & $ C \sqcap D $ & $ C^{\mathcal{I}} \cap D^{\mathcal{I}} $\\
				Existential Restriction & $ \exists R.C $ & \{ $ a \ | \ $there is $b \in \Delta^{\mathcal{I}}$ such that $ \left( a,b \right) \in R^{\mathcal{I}}$ and $  b \in C^{\mathcal{I}} $ \}\\
				Concept Subsumption& $C \sqsubseteq D$& $ C^{\mathcal{I}} \subseteq D^{\mathcal{I}} $\\				
				Role Subsumption& $R \sqsubseteq S$& $ R^{\mathcal{I}} \subseteq S^{\mathcal{I}} $\\
				Role Chain& $R_1 \circ \cdots \circ R_n \sqsubseteq R$&$ R_1^{\mathcal{I}} \circ \cdots \circ R_n^{\mathcal{I}} \subseteq R^{\mathcal{I}} $\\
				\hline
		\end{tabular}}
		\small{with $\circ$ signifying standard binary composition}\\
	\end{center}	
\end{table*}
\def\arraystretch{1}
\setlength\tabcolsep{6pt}
}

\def\completionRules{
\setlength\tabcolsep{1pt}\def\arraystretch{1.3}
\begin{table}
\caption{\ELp\ Completion Rules}
\label{tab:comprul}	\centering
	\begin{tabular}{|lccccccl|}\hline
		(1)&$A \sqsubseteq C $& &$C \sqsubseteq D$& &&$\models$&$ A \sqsubseteq D$\\
		(2)&$A \sqsubseteq C_1$& &$A \sqsubseteq C_2$& &$C_1 \sqcap C_2 \sqsubseteq D$&$\models$&$A \sqsubseteq D$\\
		(3)&$A \sqsubseteq C$& &$C \sqsubseteq \exists R.D$& &&$\models$&$A \sqsubseteq \exists R.D$\\
		(4)&$A \sqsubseteq \exists R.B$& &$B \sqsubseteq C$& &$\exists R.C \sqsubseteq D$&$\models$&$A \sqsubseteq D$\\
		(5)&$A \sqsubseteq \exists S.D$& &$S \sqsubseteq R$& &&$\models$&$A \sqsubseteq \exists R.D$\\
		(6)&$A \sqsubseteq \exists R_1.C$& &$C \sqsubseteq \exists R_2.D$& &$R_1 \circ R_2 \sqsubseteq R$&$\models$&$A \sqsubseteq \exists R.D$\\\hline
	\end{tabular}	
\end{table}
\def\arraystretch{1}\setlength\tabcolsep{6pt}
}

\def\oneSequence{
\begin{figure}
\begin{mdframed}\small
\begin{multicols}{2}
\noindent seed$_0$ = C1 $\sqsubseteq$ C2\\C2 $\sqsubseteq$ C3\\C3 $\sqcap$ C4 $\sqsubseteq$ C5\\C1 $\sqsubseteq$ $\exists$R1.C1\\C1 $\sqsubseteq$ $\exists$R2.C3\\C2 $\sqsubseteq$ $\exists$R2.C3\\C2 $\sqsubseteq$ $\exists$R1.C3\\C1 $\sqsubseteq$ $\exists$R3.C4\\C2 $\sqsubseteq$ $\exists$R1.C3\\$\exists$R1.C2 $\sqsubseteq$ C4\\R1 $\sqsubseteq$ R2\\R1 $\circ$ R3 $\sqsubseteq$ R4
\end{multicols}
Which entails
\begin{multicols}{2}
\noindent Step 1:\\C1 $\sqsubseteq$ C3\\C1 $\sqsubseteq$ C4\\C1 $\sqsubseteq$ $\exists$R1.C3\\C1 $\sqsubseteq$ $\exists$R2.C1\\C1 $\sqsubseteq$ $\exists$R4.C4
\columnbreak

\noindent Step 2:\\seed$_1$ = C1 $\sqsubseteq$ C5
\end{multicols}
\end{mdframed}
\caption{First Iteration of Sequence}
\label{fig:seq}	
\end{figure}
}

\def\piecewisePic{
\begin{figure}
    \centering
    \includegraphics[width=0.45\textwidth]{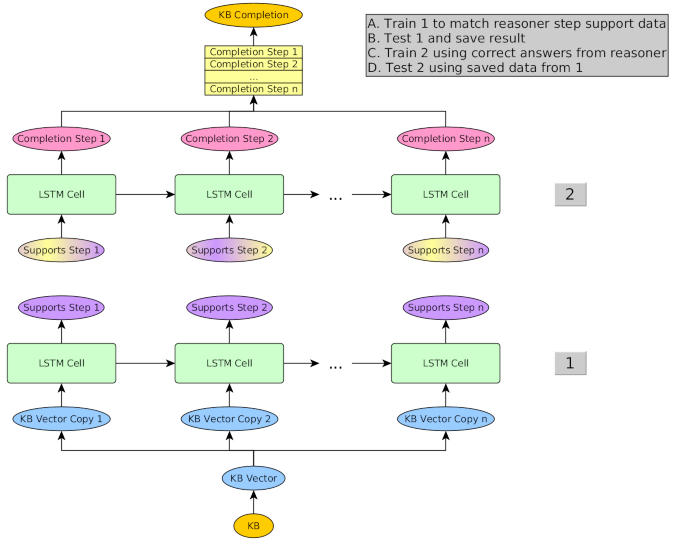}
    \caption{Piecewise Architecture}
    \label{fig:piece}
\end{figure}
}

\def\flatPic{
\begin{figure}
    \centering
    \includegraphics[width=0.35\textwidth]{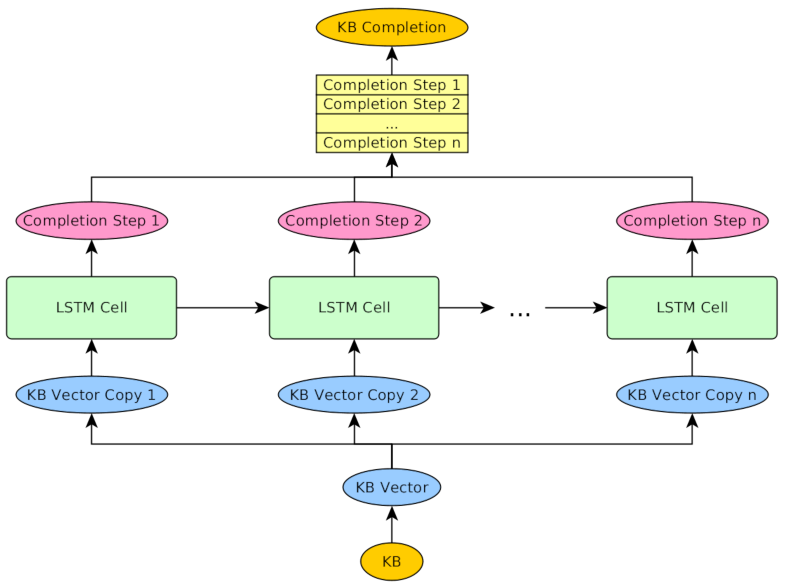}
    \caption{Flat Architecture}
    \label{fig:flat}
\end{figure}
}

\def\deepPic{
\begin{figure}
    \centering
    \includegraphics[width=0.35\textwidth]{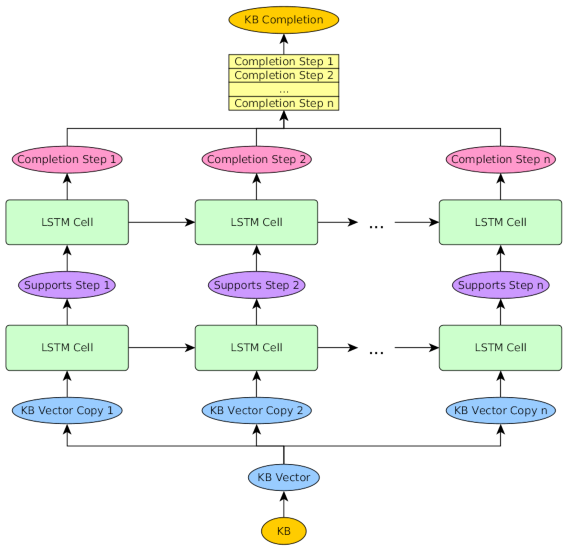}
    \caption{Deep Architecture}
    \label{fig:deep}
\end{figure}
}

\def\syntrainingPic{
\begin{figure}
    \centering
    \includegraphics[width=0.47\textwidth]{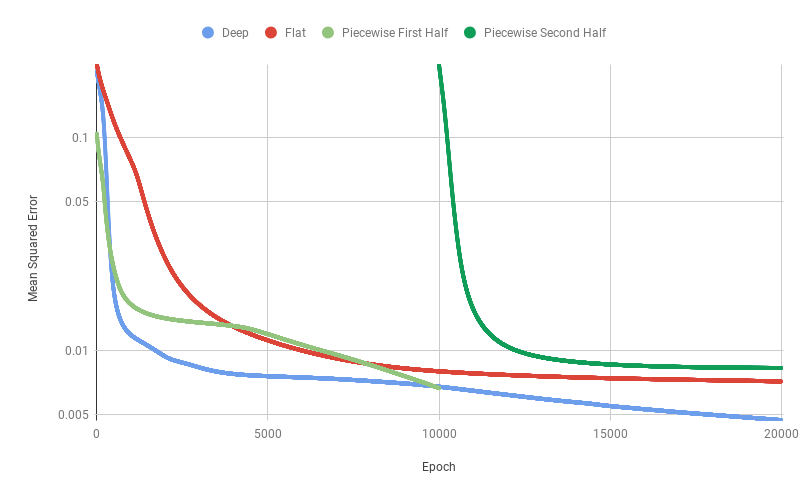}
    \caption{Synthetic Training}
    \label{fig:syntrain}
\end{figure}
}

\def\snotrainingPic{
\begin{figure}
    \centering
    \includegraphics[width=0.47\textwidth]{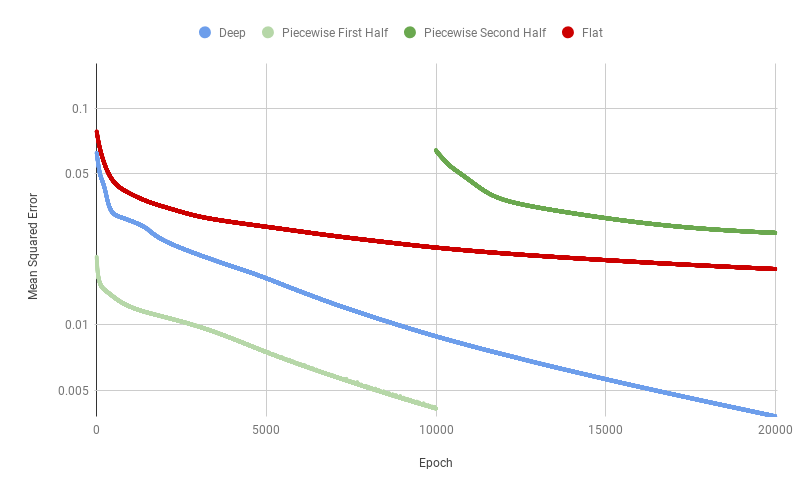}
    \caption{SNOMED Training}
    \label{fig:snotrain}
\end{figure}
}

\def\accTable{
\def\arraystretch{1.05}
\setlength\tabcolsep{.2em}
\begin{table*}[t]
\small
\caption{Average Precision Recall and F1-score For each Distance Evaluation}
\label{tab:fscore}
    \makebox[\linewidth]{
	\begin{tabular}{|l|c|c|c|c|c|c|c|c|c|}\hline
	&\multicolumn{3}{c}{Atomic Levenshtein Distance}\vline&\multicolumn{3}{c}{Character Levenshtein Distance}\vline&\multicolumn{3}{c|}{Predicate Distance}\\\cline{2-10}
	&Precision&Recall&F1-score&Precision&Recall&F1-score&Precision&Recall&F1-score\\\hline
	&\multicolumn{9}{c|}{Synthetic Data}\\\hline
	Piecewise Prediction&0.138663&0.142208&0.140412&0.138663&0.142208&0.140412&0.138646&0.141923&0.140264\\
	Deep Prediction&\textbf{0.154398}&\textbf{0.156056}&\textbf{0.155222}&\textbf{0.154398}&\textbf{0.156056}&\textbf{0.155222}&\textbf{0.154258}&\textbf{0.155736}&\textbf{0.154993}\\
	Flat Prediction&0.140410&0.142976&0.141681&0.140410&0.142976&0.141681&0.140375&0.142687&0.141521\\
	Random Prediction&0.010951&0.0200518&0.014166&0.006833&0.012401&0.008811&0.004352&0.007908&0.007908\\
	\hline
	&\multicolumn{9}{c|}{SNOMED Data}\\
	\hline	
	Piecewise Prediction&0.010530&0.013554&0.011845&0.010530&0.013554&0.011845&0.010521&0.013554&0.011839\\
	Deep Prediction&\textbf{0.015983}&0.0172811&\textbf{0.016595}&\textbf{0.015983}&0.017281&\textbf{0.016595}&\textbf{0.015614}&0.017281&\textbf{0.016396}\\
	Flat Prediction&0.014414&\textbf{0.018300}&0.016112&0.0144140&\textbf{0.018300}&0.016112&0.013495&\textbf{0.018300}&0.015525\\
	Random Prediction&0.002807&0.006803&0.003975&0.001433&0.003444&0.002023&0.001769&0.004281&0.002504\\
	\hline
	\end{tabular}}
\end{table*}
\def\arraystretch{1}
\setlength\tabcolsep{6pt}

}

\def\distTable{
\def\arraystretch{1.05}
\setlength\tabcolsep{.2em}
\begin{table*}[t]
\small
\caption{Average Statement Edit Distances with Reasoner}
\label{tab:dists}\makebox[\linewidth]{
	\begin{tabular}{|l|c|c|c|c|c|c|c|c|c|}\hline
	&\multicolumn{3}{c}{Atomic Levenshtein Distance}\vline&\multicolumn{3}{c}{Character Levenshtein Distance}\vline&\multicolumn{3}{c|}{Predicate Distance}\\\cline{2-10}
	&From&To&Average&From&To&Average&From&To&Average\\\hline
	&\multicolumn{9}{c|}{Synthetic Data}\\\hline
	Piecewise Prediction&1.336599&1.687640&1.512119&1.533115&1.812006&1.672560&2.633427&4.587382&3.610404\\
	Deep Prediction&\textbf{1.256940}&\textbf{1.507150}&\textbf{1.382045}&\textbf{1.454787}&\textbf{1.559751}&\textbf{1.507269}&\textbf{2.504496}&\textbf{3.552074}&\textbf{3.028285}\\
	Flat Prediction&1.344946&1.584674&1.464810&1.586281&1.660409&1.623345&2.517655&3.739770&3.128713\\
	Random Prediction&1.598016&1.906369&1.752192&1.970604&1.289533&1.630068&5.467918&10.57324&8.020583\\\hline
	&\multicolumn{9}{c|}{SNOMED Data}\\\hline
	Piecewise Prediction&1.704931&\textbf{2.686562}&\textbf{2.195746}&2.016249&2.862737&2.439493&6.556592&\textbf{5.857769}&6.207181\\
	Deep Prediction&1.759633&3.052080&2.405857&2.027190&3.328850&2.678020&\textbf{4.577427}&6.179389&\textbf{5.378408}\\
	Flat Prediction&\textbf{1.691738}&2.769542&2.230640&\textbf{1.948757}&2.991328&2.470042&5.548226&6.665659&6.106942\\
	Random Prediction&1.814656&3.599629&2.707143&2.094682&\textbf{1.621700}&\textbf{1.858191}&5.169093&12.392325&8.780709\\
	\hline
	\end{tabular}}
\end{table*}
\def\arraystretch{1}
\setlength\tabcolsep{6pt}
}

\def\suppExample{
\def\arraystretch{1.25}
\begin{table*}[t]
\caption{Support Generation}
\label{tab:support}	\centering\makebox[\linewidth]{
	\begin{tabular}{|l|r|c|l|}\hline
	& New Fact&Rule&Support\\\hline
	Step 1&{\color{red}C1 $\sqsubseteq$ C3}&(1)&{\color{red}C1 $\sqsubseteq$ C2,C2 $\sqsubseteq$ C3}\\
    &{\color{blue}C1 $\sqsubseteq$ C4}&(4)&{\color{blue}C1 $\sqsubseteq$ C2,C1 $\sqsubseteq \exists$R1.C1,$\exists$R1.C2 $\sqsubseteq$ C4}\\
    &C1 $\sqsubseteq \exists$R1.C3&(3)&C1 $\sqsubseteq$ C2,C2 $\sqsubseteq \exists$R1.C3\\
    &C1 $\sqsubseteq \exists$R2.C1&(5)&C1 $\sqsubseteq \exists$R1.C1,R1 $\sqsubseteq$ R2\\
    &C1 $\sqsubseteq \exists$R4.C4&(6)&C1 $\sqsubseteq \exists$R1.C1,R1 $\circ$ R3 $\sqsubseteq$ R4,C1 $\sqsubseteq \exists$R3.C4\\\hline
    Step 2&C1 $\sqsubseteq$ C5&(2)&C3 $\sqcap$ C4 $\sqsubseteq$ C5,{\color{red}C1 $\sqsubseteq$ C2,C2 $\sqsubseteq$ C3},{\color{blue}C1 $\sqsubseteq$ C2,C1 $\sqsubseteq \exists$R1.C1,$\exists$R1.C2 $\sqsubseteq$ C4}\\\hline
	\end{tabular}}	
\end{table*}
\def\arraystretch{1}
}

\def\exampleSynPred{
\def\arraystretch{1.25}
\begin{table}
\caption{Example Synthetic Output}
\label{tab:synex}	\centering\makebox[\linewidth]{
	\begin{tabular}{|l|l|l|}\hline
	&Correct Answer&Predicted Answer\\\hline
	Step 0&C9 $\sqsubseteq$ C11&C8 $\sqsubseteq$ C9\\
	&C2 $\sqsubseteq$ C10&C1 $\sqsubseteq$ C9\\
	&C9 $\sqsubseteq$ C12&C8 $\sqsubseteq$ C9\\
	&C7 $\sqsubseteq$ C6&C8 $\sqsubseteq$ $\exists$R4.C9\\
	&C9 $\sqsubseteq$ $\exists$R4.C11&C1 $\sqsubseteq$ $\exists$R5.C9\\
	&C2 $\sqsubseteq$ $\exists$R4.C9&C8 $\sqsubseteq$ $\exists$R4.C9\\
	&C9 $\sqsubseteq$ $\exists$R5.C9&C9 $\sqsubseteq$ $\exists$R5.C9\\
	&C2 $\sqsubseteq$ $\exists$R5.C11&\\
	&C9 $\sqsubseteq$ $\exists$R7.C12&\\
	&C2 $\sqsubseteq$ $\exists$R6.C12&\\\hline
	Step 1&C9 $\sqsubseteq$ C13&C8 $\sqsubseteq$ C12\\
	&C2 $\sqsubseteq$ C11&C2 $\sqsubseteq$ C10\\
	&C2 $\sqsubseteq$ C12&C1 $\sqsubseteq$ C11\\
	&C2 $\sqsubseteq$ $\exists$R4.C11&C1 $\sqsubseteq$ $\exists$R3.C12\\
	&C2 $\sqsubseteq$ $\exists$R5.C9&C1 $\sqsubseteq$ $\exists$R4.C8\\
	&C2 $\sqsubseteq$ $\exists$R7.C12&\\\hline		
	Step 2&C2 $\sqsubseteq$ C13&C1 $\sqsubseteq$ C12\\	
	\hline
	\end{tabular}}	
\end{table}
\def\arraystretch{1}
}

\def\exampleSnoPred{
\begin{table}
\caption{Example SNOMED Outputs}
\label{tab:snoex}	\centering\makebox[\linewidth]{
	\begin{tabular}{|l|l|}\hline
Correct Answer&C6 $\sqsubseteq$ $\exists$R4.C1\\\hline
Meaning&if something is a zone of superficial fascia, then\\
&there is a subcutaneous tissue that it is PartOf\\\hline\hline
Prediction&C6 $\sqsubseteq$ $\exists$R4.C3\\\hline
Meaning&if something is a zone of superficial fascia,\\
&then there is a subcutaneous tissue of palmar\\
&area of little finger that it is PartOf\\\hline\hline
Correct Answer&C8 $\sqsubseteq$ $\exists$R3.C2\\\hline
Meaning&if something is a infrapubic region of pelvis,\\
&then there is a perineum that it is PartOf\\\hline\hline
Prediction&C9 $\sqsubseteq$ $\exists$R3.C2\\\hline
Meaning&if something is a zone of integument,\\
&then there is a perineum that it is PartOf\\
\hline
	\end{tabular}}	
\end{table}
}

\def\goals{
\begin{enumerate}
    \item Can a neural network emulate reasoner behavior?
    \item Can we still achieve success without embeddings?
    \item Can learning happen with arbitrary numerical names?
    \item Will intermediate answers help our evaluation?
    \item Is F1-score sufficient to evaluate an integrated logic and neural network system? If not, what would be?
\end{enumerate}
}

\def\predAcc{
\begin{figure*}%
    \centering
    \subfloat[Synthetic Data Piecewise Architecture]{{\includegraphics[width=0.32\textwidth]{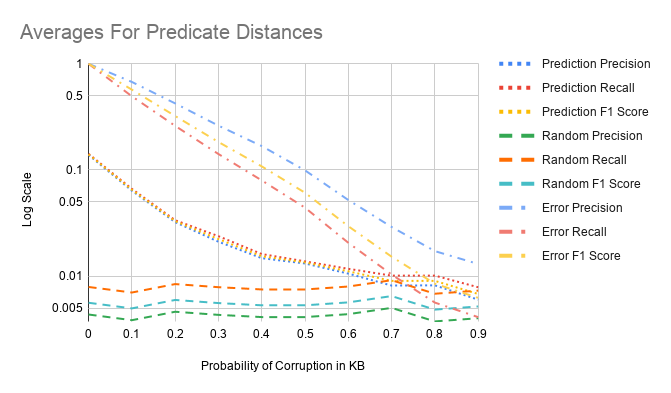}}}%
    \subfloat[Synthetic Data Deep Architecture]{{\includegraphics[width=0.32\textwidth]{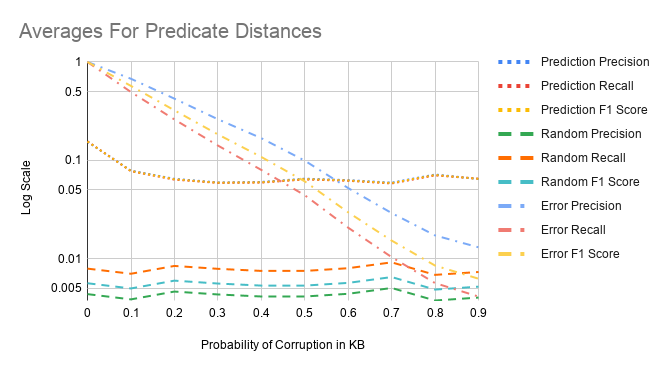}}}%
    \subfloat[Synthetic Data Flat Architecture]{{\includegraphics[width=0.32\textwidth]{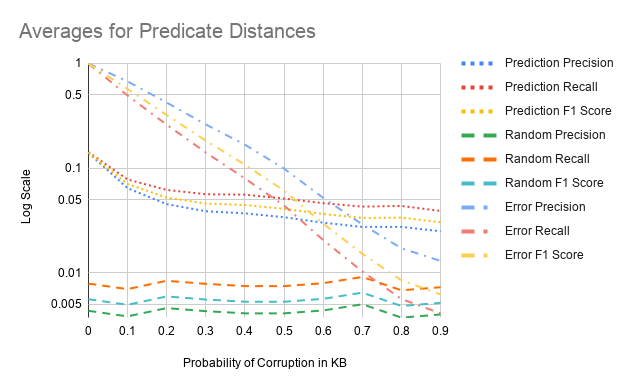}}}%
    \qquad
    \subfloat[SNOMED Data Piecewise Architecture]{{\includegraphics[width=0.32\textwidth]{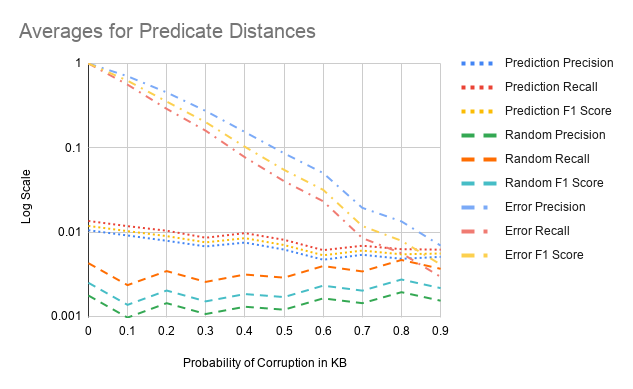}}}%
    \subfloat[SNOMED Data Deep Architecture]{{\includegraphics[width=0.32\textwidth]{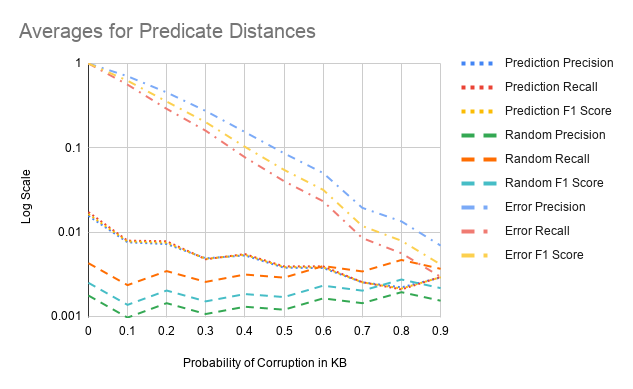}}}%
    \subfloat[SNOMED Data Flat Architecture]{{\includegraphics[width=0.32\textwidth]{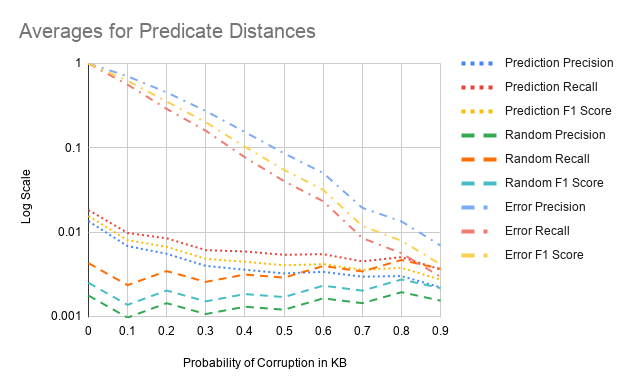}}}%
    \caption{Predicate Distance Precision, Recall, and F1-score}%
    \label{fig:fpred}%
\end{figure*}
}

\def\levAcc{
\begin{figure*}%
    \centering
    \subfloat[Synthetic Data Piecewise Architecture]{{\includegraphics[width=0.32\textwidth]{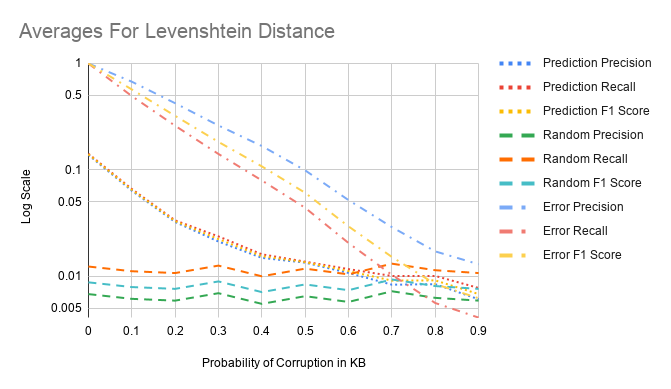}}}%
    \subfloat[Synthetic Data Deep Architecture]{{\includegraphics[width=0.32\textwidth]{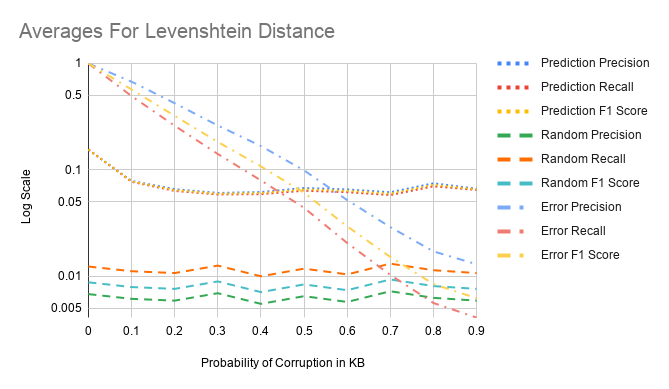}}}%
    \subfloat[Synthetic Data Flat Architecture]{{\includegraphics[width=0.32\textwidth]{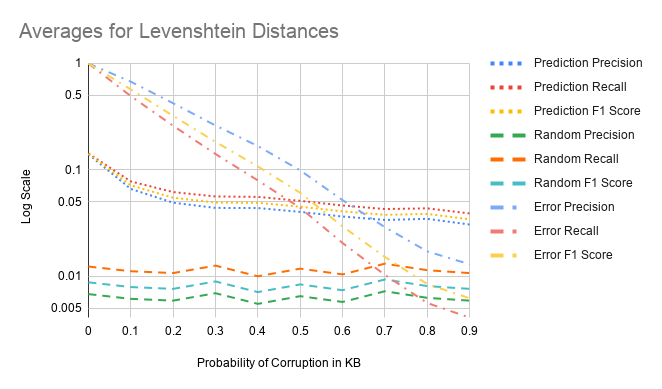}}}%
    \qquad
    \subfloat[SNOMED Data Piecewise Architecture]{{\includegraphics[width=0.32\textwidth]{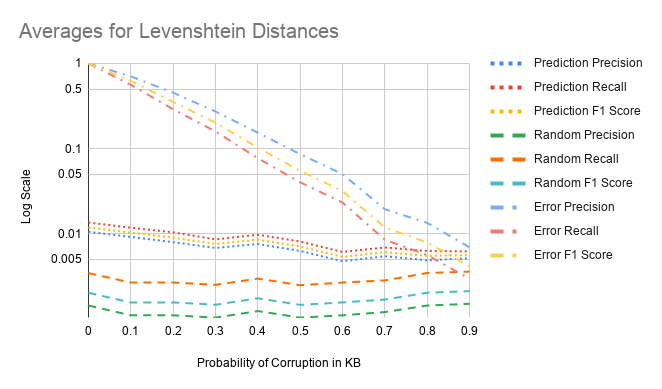}}}%
    \subfloat[SNOMED Data Deep Architecture]{{\includegraphics[width=0.32\textwidth]{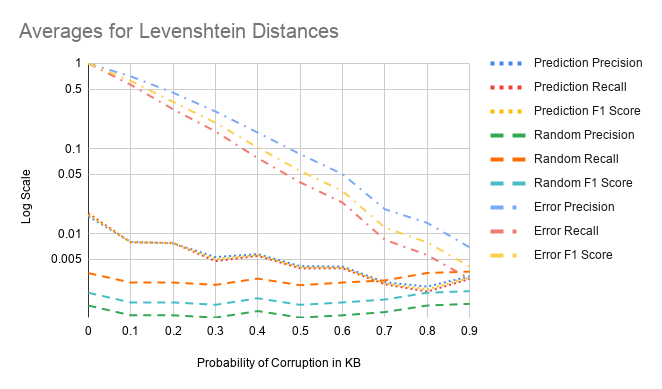}}}%
    \subfloat[SNOMED Data Flat Architecture]{{\includegraphics[width=0.32\textwidth]{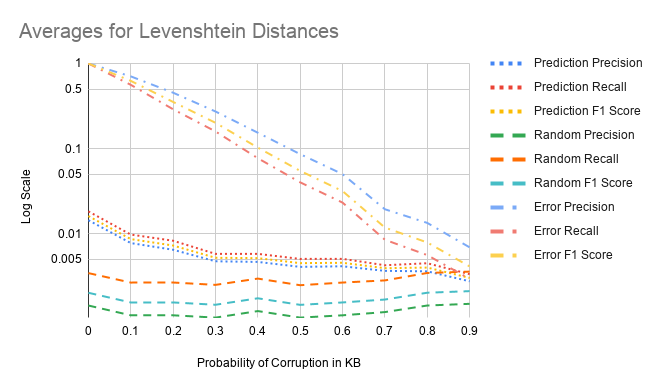}}}%
    \caption{Character Levenshtein Distance Precision, Recall, and F1-score}%
    \label{fig:flev}%
\end{figure*}
}

\def\atomAcc{
\begin{figure*}%
    \centering
    \subfloat[Synthetic Data Piecewise Architecture]{{\includegraphics[width=0.32\textwidth]{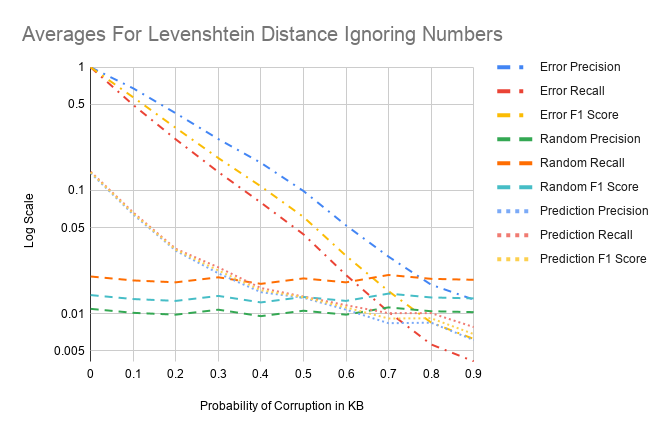}}}%
    \subfloat[Synthetic Data Deep Architecture]{{\includegraphics[width=0.32\textwidth]{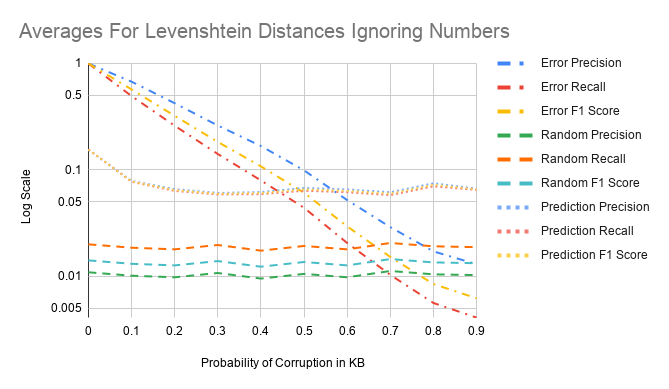}}}%
    \subfloat[Synthetic Data Flat Architecture]{{\includegraphics[width=0.32\textwidth]{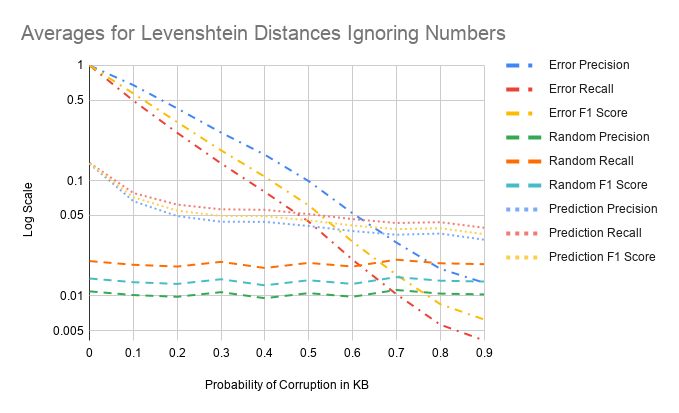}}}%
    \qquad
    \subfloat[SNOMED Data Piecewise Architecture]{{\includegraphics[width=0.32\textwidth]{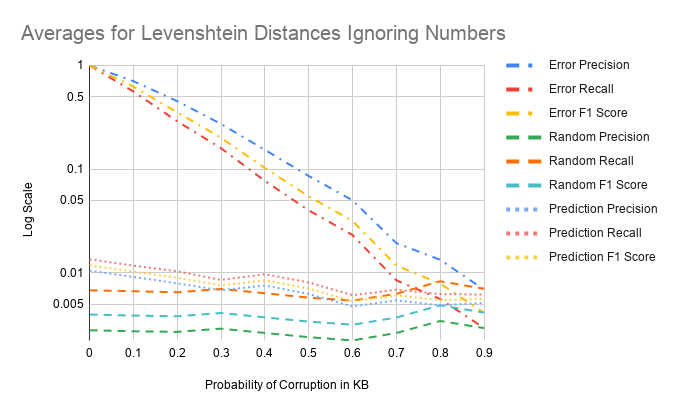}}}%
    \subfloat[SNOMED Data Deep Architecture]{{\includegraphics[width=0.32\textwidth]{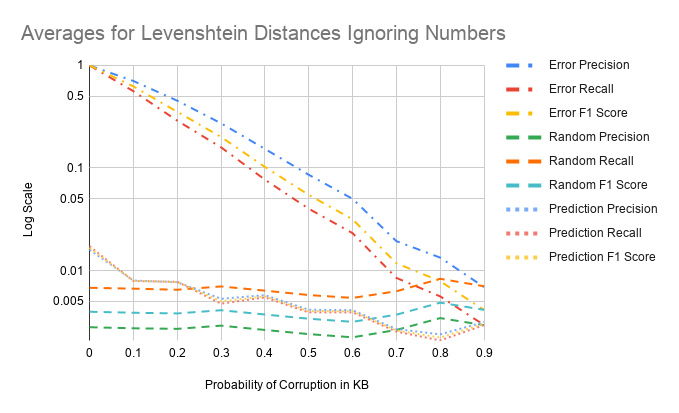}}}%
    \subfloat[SNOMED Data Flat Architecture]{{\includegraphics[width=0.32\textwidth]{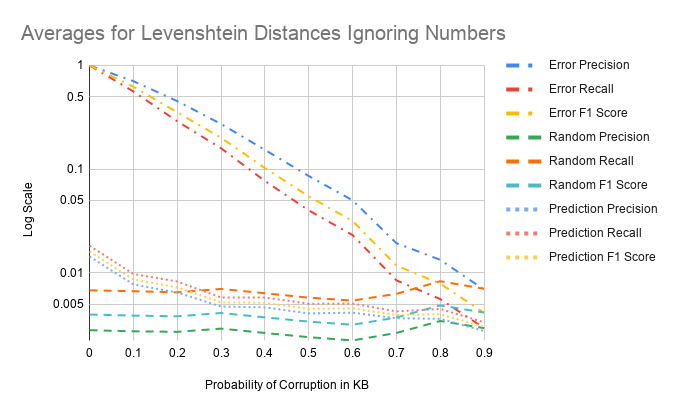}}}%
    \caption{Atomic Levenshtein Distance Precision, Recall, and F1-score}%
    \label{fig:fatom}%
\end{figure*}
}

\def\predDist{
\begin{figure*}%
    \centering
    \subfloat[Synthetic Data Piecewise Architecture]{{\includegraphics[width=0.32\textwidth]{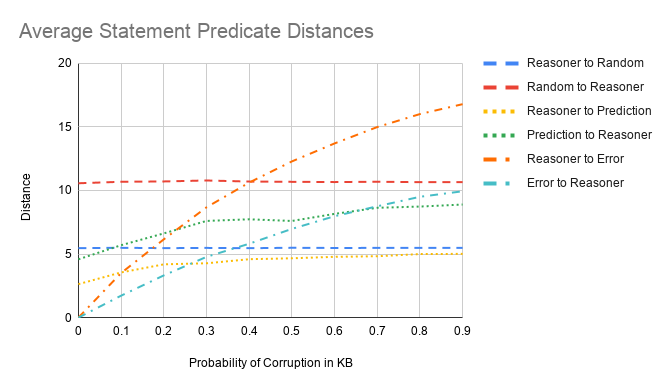}}}%
    \subfloat[Synthetic Data Deep Architecture]{{\includegraphics[width=0.32\textwidth]{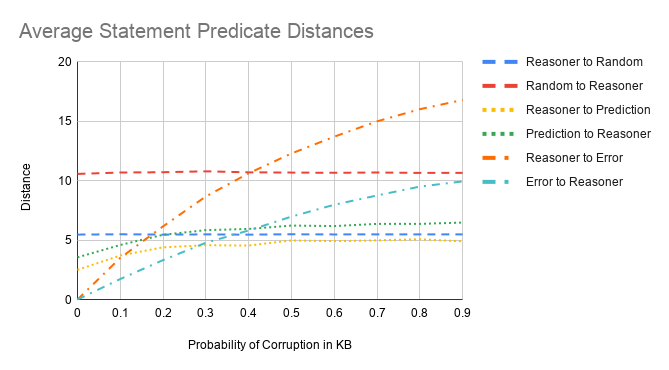}}}%
    \subfloat[Synthetic Data Flat Architecture]{{\includegraphics[width=0.32\textwidth]{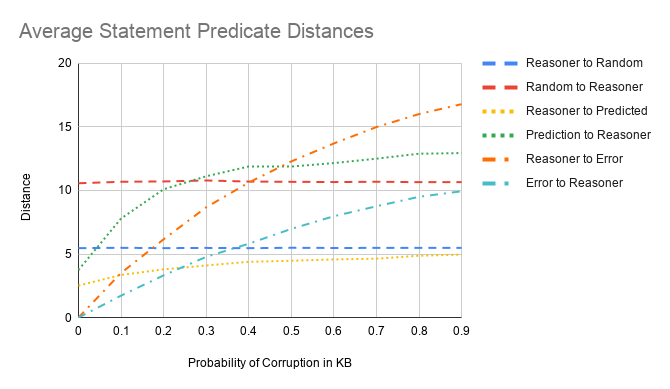}}}%
    \qquad
    \subfloat[SNOMED Data Piecewise Architecture]{{\includegraphics[width=0.32\textwidth]{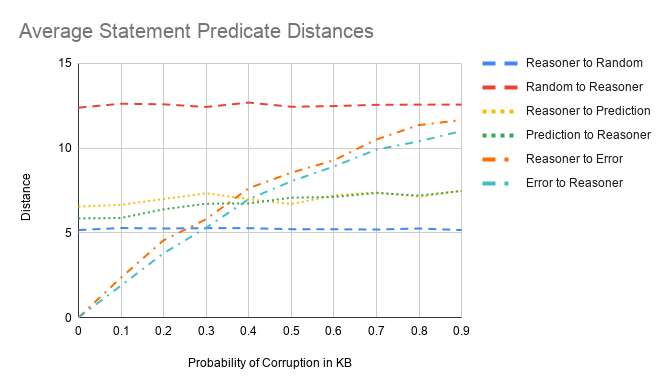}}}%
    \subfloat[SNOMED Data Deep Architecture]{{\includegraphics[width=0.32\textwidth]{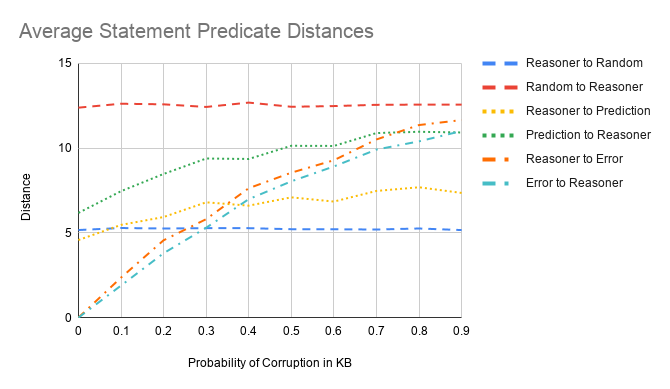}}}%
    \subfloat[SNOMED Data Flat Architecture]{{\includegraphics[width=0.32\textwidth]{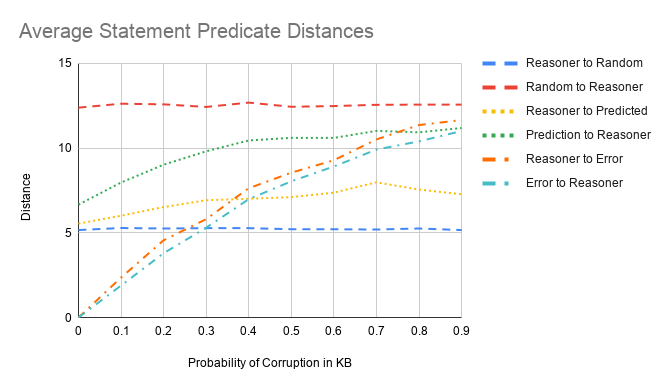}}}%
    \caption{Predicate Distance}%
    \label{fig:distpred}%
\end{figure*}
}

\def\levDist{
\begin{figure*}%
    \centering
    \subfloat[Synthetic Data Piecewise Architecture]{{\includegraphics[width=0.32\textwidth]{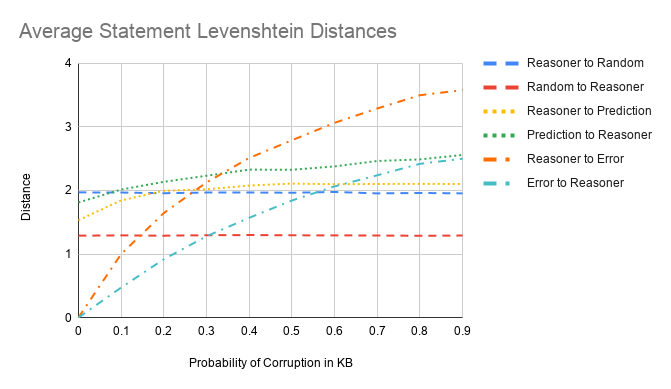}}}%
    \subfloat[Synthetic Data Deep Architecture]{{\includegraphics[width=0.32\textwidth]{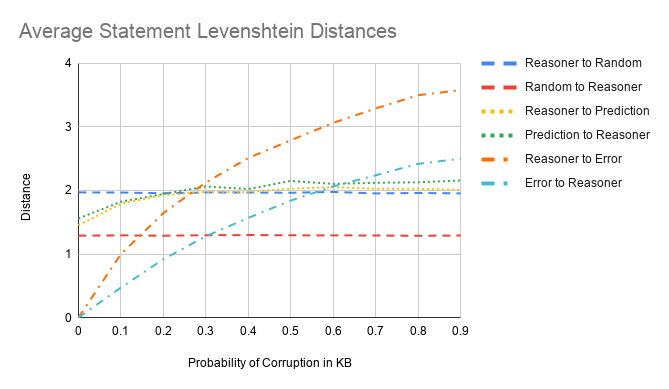}}}%
    \subfloat[Synthetic Data Flat Architecture]{{\includegraphics[width=0.32\textwidth]{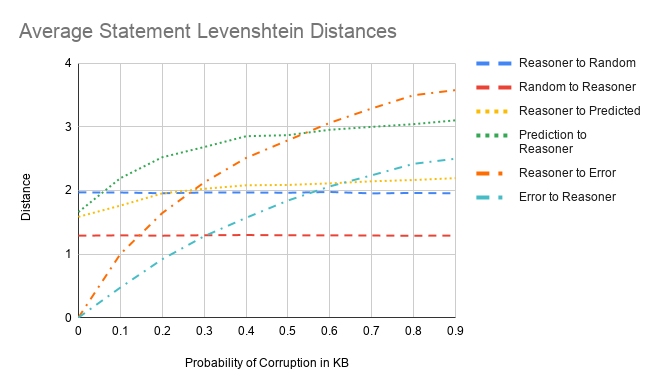}}}%
    \qquad
    \subfloat[SNOMED Data Piecewise Architecture]{{\includegraphics[width=0.32\textwidth]{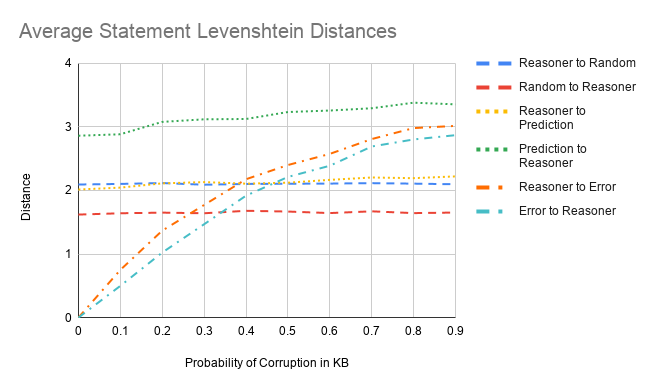}}}%
    \subfloat[SNOMED Data Deep Architecture]{{\includegraphics[width=0.32\textwidth]{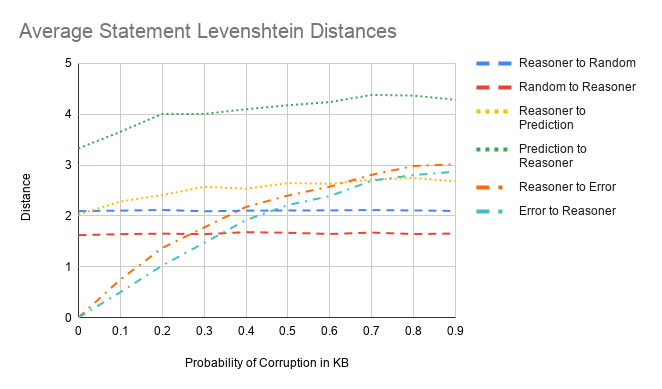}}}%
    \subfloat[SNOMED Data Flat Architecture]{{\includegraphics[width=0.32\textwidth]{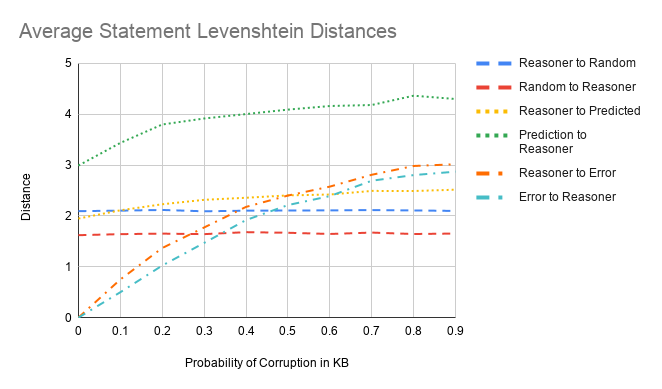}}}%
    \caption{Character Levenshtein Distance}%
    \label{fig:distlev}%
\end{figure*}
}

\def\atomDist{
\begin{figure*}%
    \centering
    \subfloat[Synthetic Data Piecewise Architecture]{{\includegraphics[width=0.32\textwidth]{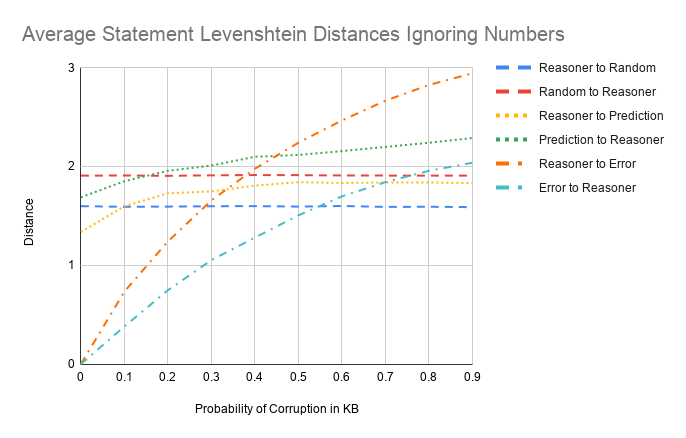}}}%
    \subfloat[Synthetic Data Deep Architecture]{{\includegraphics[width=0.32\textwidth]{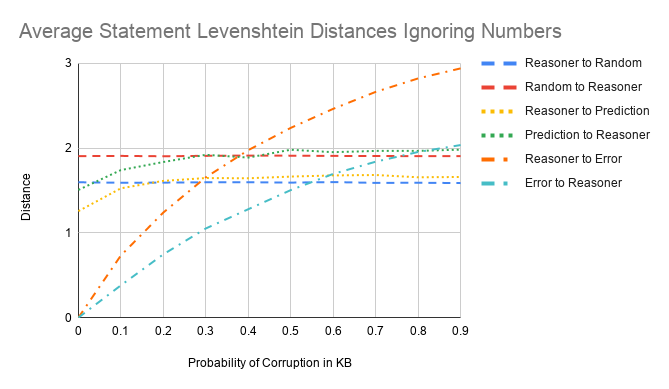}}}%
    \subfloat[Synthetic Data Flat Architecture]{{\includegraphics[width=0.32\textwidth]{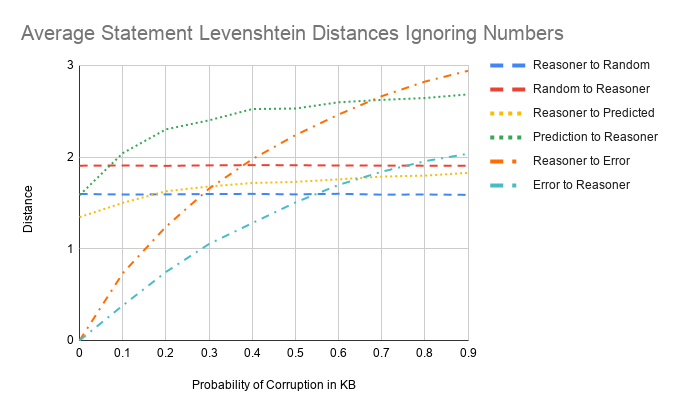}}}%
    \qquad
    \subfloat[SNOMED Data Piecewise Architecture]{{\includegraphics[width=0.32\textwidth]{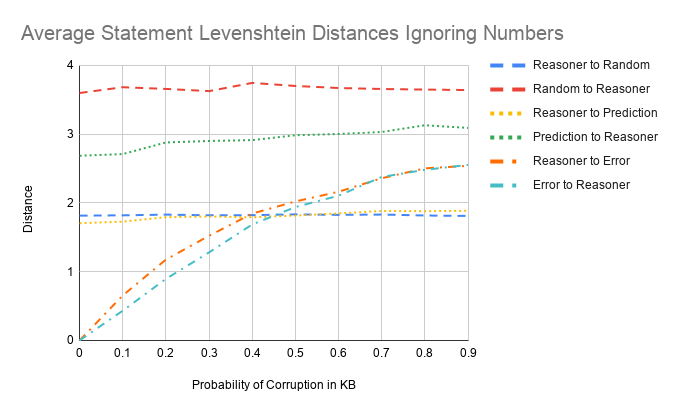}}}%
    \subfloat[SNOMED Data Deep Architecture]{{\includegraphics[width=0.32\textwidth]{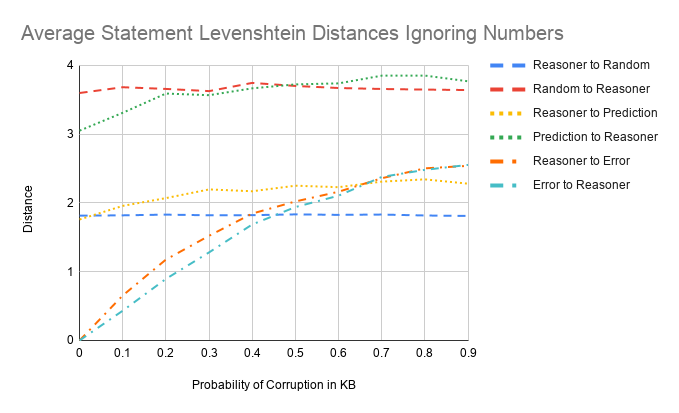}}}%
    \subfloat[SNOMED Data Flat Architecture]{{\includegraphics[width=0.32\textwidth]{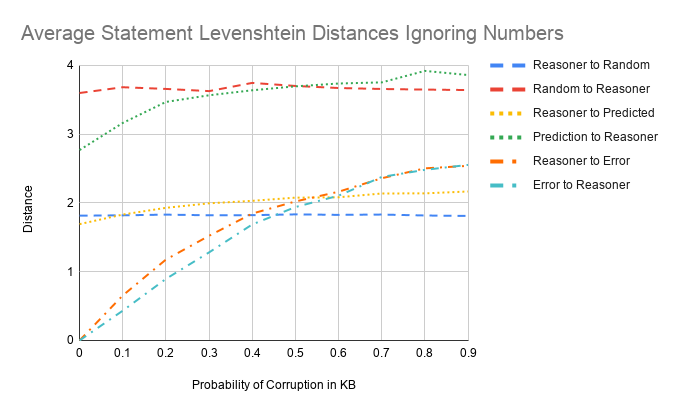}}}%
    \caption{Atomic Levenshtein Distance}%
    \label{fig:distatom}%
\end{figure*}
}

%% file: macros.tex
\def\ELp{$\mathcal{EL}^+$}

\def\ELpp{$\mathcal{EL}^{++}$}

\def\nls{\par\null\par}

\newcommand{\sect}[1]{\section{#1}\label{sec:#1}}

\newcommand{\subsect}[1]{\subsection{#1}\label{subsec:#1}}

\newcommand{\subsubsect}[1]{\subsubsection{#1}\label{subsubsec:#1}}